\documentclass{article}

\usepackage{arxiv}

\usepackage[utf8]{inputenc} 
\usepackage[T1]{fontenc}    
\usepackage{hyperref}       
\usepackage{url}            
\usepackage{booktabs}       
\usepackage{amsfonts}       
\usepackage{nicefrac}       
\usepackage{microtype}      
\usepackage{lipsum}		
\usepackage{graphicx}
\usepackage{natbib}
\usepackage{doi}
\usepackage{xcolor}
\usepackage{bm}
\usepackage{amsthm}
\usepackage{multirow}
\usepackage{siunitx}
\usepackage{subcaption}
\usepackage{amsmath}

\usepackage{array}
\newcolumntype{P}[1]{>{\centering\arraybackslash}p{#1}}

\newcommand*{\etc}{%
    \@ifnextchar{.}%
        {etc}%
        {etc.}%
}

\definecolor{ao}{rgb}{0.0, 0.5, 0.0}

\newtheorem{theorem}{Theorem}

\newtheorem{lemma}[theorem]{Lemma}
\newtheorem{corollary}[theorem]{Corollary}
\newtheorem{definition}{Definition}

\hypersetup{
    colorlinks=true,
    urlcolor=cyan,
    citecolor=magenta,
    }

\title{
Constrained Monotonic Neural Networks}

\author{ \href{https://orcid.org/0000-0001-6912-9900}{\includegraphics[scale=0.06]{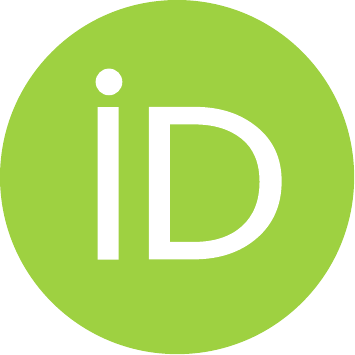}\hspace{1mm}Davor Runje}\thanks{AIRT Reasearch and Algebra University College} \\
	\url{davor@airt.ai} \\
	\And
	\href{https://orcid.org/0000-0003-3482-4887}{\includegraphics[scale=0.06]{orcid.pdf}\hspace{1mm}Sharath M. Shankaranarayana} \thanks{AIRT Reasearch}\\
	\url{sharath@airt.ai} \\
}

\hypersetup{
pdftitle={Constrained Monotonic Neural Networks},
pdfsubject={cs.ML, cs.AI},
pdfauthor={Davor Runje, Sharath M. Shankaranarayana},
pdfkeywords={monotonicity, deep learning},
}

\begin{document}
\maketitle

\begin{abstract}
	
	
	
	Deep neural networks are becoming increasingly popular in approximating arbitrary functions from noisy data.  But wider adoption is being hindered by the need to explain such models and to impose additional constraints on them. Monotonicity constraint is one of the most requested properties in real-world scenarios and is the focus of this paper.  
	One of the oldest ways to construct a monotonic fully connected neural network is to constrain its weights to be non-negative while employing a monotonic activation function. Unfortunately, this construction does not work with popular non-saturated activation functions such as ReLU, ELU, SELU etc, as it can only approximate convex functions. We show this shortcoming can be fixed by employing the original activation function for a part of the neurons in the layer, and employing its point reflection for the other part. Our experiments show this approach of building monotonic deep neural networks have matching or better accuracy when compared to other state-of-the-art methods such as deep lattice networks or monotonic networks obtained by heuristic regularization. This method is the simplest one in the sense of having the least number of parameters, not requiring any modifications to the learning procedure or  steps post-learning steps. Finally, we show how this approach could be extended to other architectures such as convolutional neural networks and recurrent neural networks. 
	
\end{abstract}

\keywords{monotonicity \and deep learning}

\section{Introduction}
Deep Learning has witnessed widespread adoption in many critical real-world domains such as finance, healthcare, etc \citep{LeCun2015}. Predictive models built using deep neural networks have been shown to have high accuracy. 
Incorporating prior knowledge such as monotonicity in trained models help in improving the performance and generalization ability of the trained models \citep{Mitchell80theneed,dugas2000incorporating}. Introduction of structural biases such as monotonicity make models also more data-efficient, enabling a leap in predictive power on smaller datasets \citep{velickovic2019phd}.
Apart from the requirements of having models with high accuracy, there is also a need for transparency and interpretability and monotonicity helps in partially achieving the above requirements \citep{gupta2016monotonic}. 
Due to legal, ethical and/or safety concerns, monotonicity of predictive models with respect to some input or all the inputs is required in numerous domains such as financial (house pricing, credit scoring, insurance risk), healthcare (medical diagnosis, patient medication) and legal (criminal sentencing) to list just a few. All other things being equal, a larger house should be deemed more valuable, a bank's clients with higher income should be eligible for a larger loan \citep{sharma2020higher}, an offender with a longer crime history should be predicted as more likely to commit another crime \citep{Rudin2020Age}, etc. A model without such a monotonic property would not, and certainly should not, be trusted by the society to provide a basis for such important decisions. However, the monotonicity of deep learning models is not a guaranteed property even when trained on monotonic data, let alone when training on noisy data typically used in practice.

Although, monotonicity is an important and increasingly often even required property, there is no simple or easy method to enforce it. It has been an active area of research and the existing methods on the subject can be broadly categorized into two types:

\begin{enumerate}
    \item Monotonic architectures by construction: This line of research focuses on neural architectures guaranteeing monotonicity by construction \citep{archer1993application,sill1997monotonic,daniels2010monotone, milani2016fast, you2017deep}.

    %

    \item Monotonicity by regularization:  This line of research focuses on enforcing monotonicity in neural networks during training by employing a modified loss function or a heuristic regularization term \citep{sill1996monotonicity,gupta2019incorporate}. However, employing a modified loss function or a regularization term give us no guarantees that the trained model is indeed monotonic. A recent line of research work focuses on verifying/certifying monotonicity of such models \citep{sivaraman2020counterexample, liu2020certified}. These approaches are computationally very expensive and they require multiple retraining cycles to enforce monotonicity.
    
\end{enumerate}
We will give a more detail account of the existing methods in the next section.

The simplest method to achieve monotonicity by construction is to constrain the weights of the fully connected neural network to have only non-negative (for non-decreasing variables) or only non-positive values (for non-ascending) variables when used in conjunction with a monotonic activation function, a technique known for almost 30 years \citep{archer1993application}. However, this method does not work well in practice. When used in conjuction with saturated (bounded) activation functions such as the sigmoid  and hyperbolic tangent, these models are difficult to train, i.e. they do not converge to a good solution. On the other hand, when used with non-saturated (unbounded) convex activation functions such as ReLU \citep{nair2010rectified}, the resulting models are always convex \citep{liu2020certified}, severely limiting the applicability of the method in practice.

The main contribution of this paper is a simple modification of the method above which, in conjuction with non-saturated activation functions, is capable of approximating non-convex functions as well: if both concave and convex monotone activation functions are used in a neural network with constrained weights, it regains the ability to approximate monotone continuous functions that are either convex, concave or nothing of the above.


The resulting model is guaranteed to be monotonic, can be used in conjunction with any convex monotonic non-saturated activation function, doesn't have any additional parameters compared to a non-monotonic fully-connected network for the same task and can be trained without any additional requirements on the learning procedure. Experimental results show it is exceeding the performance of all other state-of-the-art methods, all while being both simpler (in number of parameters) and easier to train.

 




Our contributions can be summarized as follows:
\begin{enumerate}
    \item We propose a modification to an existing constricted neural network layer enabling it to model non-convex functions when used with non-saturated monotone convex activation functions such as ReLU, ELU, SELU and alike.
    
    \item We prove monotonicity property of neural architectures using the proposed constricted neural network layer and ReLU activation function.
    
    \item We perform extensive ablation studies to show empirically as to how our proposed monotonic dense layer performs.
    
    \item We perform comparison with other recent works and show that our proposed novel neural network block can yield comparable and in some cases better results than the previous state-of-the-art and with significantly less parameters.
\end{enumerate}

\section{Related work}

Methods to build monotonic models can be broadly categorized into two broad categories: monotonic by construction and monotonic by regularization. Before dwelling deeper into those methods, we give a short overview of activation functions as they are crucial to understanding our work. Finally, we discuss universal approximation theorem for different kind of neural architectures.

\subsection{Activation functions}

Right from it's inception in perceptron \citep{rosenblatt1958perceptron}, non linear activation functions have historically been one of the most important components of neural networks. Previously, the saturated functions such as the sigmoid \citep{rumelhart1986learning}, the hyperbolic tangent \citep{neal1992connectionist} and its variants were the most common choice of activation functions. 

Currently, one of the most important factors for state-of-the-art results accomplished by the modern neural networks is use of non-saturated activation functions. The use of \emph{Rectified Linear Unit} (ReLU) \citep{nair2010rectified, glorot2011deep} as activation function was instrumental in achieving good performance in newer architectures. The ReLU has since become defacto choice of activation in most of the practical implementations and continues to be widely used because of its advantages such as simple computation, representational sparsity and linearity. 

Later, small modifications to ReLU activation function termed \emph{Leaky ReLU} \citep{maas2013rectifier} and \emph{Exponential Linear Unit} (ELU) \citep{clevert2015fast} were proposed to address the problem of dead neurons in ReLU. A parametic form of Leaky ReLU termed \emph{Parametric ReLU} (PReLU) was later proposed in \citep{he2015delving}. The authors of \cite{he2015delving} also introduced a better initialization scheme for these rectifier-like activation functions that helped in exploring powerful deep neural architectures and was one of the first works that surpassed the human performance in Imagenet classification task \citep{russakovsky2015imagenet}.
The major advantages attributed to the use of non-saturated activation functions such as ReLU, Leaky ReLU, ELU and PReLU over the saturated ones such as \emph{sigmoid} and \emph{hyperbolic tangent} are alleviation of the problem of exploding and vanishing gradient and 
acceleration of the convergence during training \citep{xu2015empirical}.

The more recent works on activation functions such as \emph{Softplus} \citep{zheng2015improving}, \emph{Gaussian Error Linear Units} (GELU) \citep{hendrycks2016gaussian}, \emph{Swish} \citep{ramachandran2017swish}, \emph{Scaled Exponential Linear Unit} (SELU) \citep{klambauer2017self} etc., have broadly focussed on solving problems of dead ReLU and aid in faster convergence. These activation functions have been shown to provide improved results and hence tend to be useful in practice. 

ReLU and its variants LeakyReLU, PReLU, ELU, SELU and Softmax have the following important properties for our work: they are non-decreasing, monotonic and convex. Notice that this is not a general rule for activation functions used today, e.g. GELU, Swish are not monotonic and we will not use it in our construction.

The idea of using the both the original activation function and its point reflection in the same layer has been proposed in \cite{DBLP:journals/corr/ShangSAL16} where both outputs of ReLU and the negative value of its point reflection were used in construction of \emph{concatenated ReLU} (CReLU) activation function. The proposed modification outputs two values instead of one and therefore increases the number of parameters. In \cite{Zagoruyko2017diracnets}, the authors propose \emph{negative concatenated ReLU} (NCReLU) flip the sign and use the point reflection directly. Notice that proposed architectural change could be applied to other non-saturated, monotonic activation functions as well, but with unknown impact on their performance.

In \cite{DBLP:conf/iclr/EidnesN18}, the authors propose \emph{bipolar ReLU} which consists of using ReLU on the half of the neurons in the layer and the point reflection of ReLU on the other half. The same construction could be used with other ReLU family activation functions as well. However, the focus of their work was to alleviate the need of normalizing layers and improving the performance of deep vanilla recurrent neural networks (RNNs) by using \emph{bipolar ReLU}.

Both NCReLU and bipolar ReLU could have been be used in construction of constrained neural network capable of representing non-convex function, but to the best of our knowledge they have not been so far. The focus of attention when they were introduced was to increase performance of CNN networks on computer vision tasks and that was probably the reason why this, although important application, was overlooked.



\subsection{Monotonicity by construction}

Provided a monotone, non-decreasing activation function, a straightforward method to enforce non-decreasing monotonicity of a fully connected neural network is to constrain its weights to be non-negative as proposed in \cite{archer1993application}.
However, when compared to a neural network with a non-saturating activation function, the neural network using a saturating activation function (such as sigmoid and hyperbolic tangent) obtains inferior performance as shown in \cite{nair2010rectified, glorot2011deep}. This is also true in the case of constrained neural networks as shown by the experiments in the Section \ref{sec:experiments}. 

If we replace sigmoid with a non-decreasing monotone convex activation function such as ReLU, resulting networks are easy to train and but can approximate convex functions only \citep{liu2020certified}. This limitation makes the method highly impractical as in the real-world use cases the functions modelled are rarely convex.


Another approach to build monotonic neural architecture are Min-Max networks where monotonic linear embedding and max-min-pooling are used \citep{sill1997monotonic}. In \cite{daniels2010monotone}, authors generalized this approach to handle functions that are partially monotonic and proved that the resulting networks have the universal approximation property. However, such networks are very difficult to train and not used in practice. Their construction does not allow for replacement with other activation functions.


Deep lattice networks (DLN) \citep{you2017deep} use a combination of linear calibrators and lattices \citep{milani2016fast} for learning monotonic functions. This is the most widely used method in practice today, but not without its limitation. Lattices are structurally rigid thereby restricting the hypothesis space significantly. Also, DLN requires a very large number of parameters to obtain good performance.

\subsection{Monotonicity by regularization}
In the second category, the research works focus on enforning monotonicity during the training process by modifying the
loss function or by adding additional regularization term. 

In \cite{sill1996monotonicity}, the authors propose a modified loss function
that penalizes non-monotonicity of the model. The algorithm models the input distribution as a joint Gaussian estimated from the training data and samples random pairs of monotonic points that are added to the training data. In \cite{gupta2019incorporate}, the authors propose a point-wise loss function that acts as a soft monotonicity constraint. These methods are straightforward to implement and can be used with any neural network architecture. However, these methods do not guarantee monotonicity of trained model. 

Recently, there is an increasing number of proposed methods to certify or verify monotonicity obtained by regularization methods. In  \cite{liu2020certified}, the authors propose an optimization-based technique for mathematically verifying, or rejecting, the
monotonicity of an arbitrary piece-wise linear (e.g., ReLU) neural network. The method consists of transforming the monotonicity verification problem to a mixed integer linear programming (MILP) problem that can be solved using an off-the-shelf MILP solver.

In \cite{sivaraman2020counterexample}, the authors propose an approach that finds counterexamples (defined as the pair of points where the monotonicity constraint
is violated) by employing satisfiability modulo theories (SMT) solver \citep{barrett2018satisfiability}. To satisfy the monotonicity constraints, these counterexamples are included in the training data with adjustments to their target values so as to enforce next iterations of the model to be monotonic.

Both methods \cite{liu2020certified,sivaraman2020counterexample} have been shown to supports ReLU as the activation function only and there is no obvious way how to extend them to other activation functions. More precisely, they rely on piece-wise linearity of ReLU to work, the property not satisfied by other variants such as ELU, SELU, GELU, etc. Last but not least, the procedure for certifying/verifying using MILP or SMT solvers is computationally very costly. These approaches also require multiple reruns or iterations to arrive at certified/verified monotonic networks.

\subsection{Universal approximation property}

The classical Universal Approximation Theorem \citep{Cybenko1989,hornik1991,pinkus1999} states that any continuous function on a closed interval can be approximated with a feed-forward neural network with one hidden layer if and only if its activation function is nonpolynomial. In \cite{kidger2020}, authors prove the approximation property holds for arbitrary \emph{deep} neural networks with bounded number of neurons in each layer holds if the activation function is nonaffine and differential at at least one point.

The universal approximation property for constrained multivariate neural networks using \emph{sigmoid} as the activation functions states that any multivariate continuous monotone function on a compact subset of $\mathbb{R}^k$ can be approximated with a constrained neural network of at most $k$ layers \cite{daniels2010monotone}. 

However, the universal approximation theorem does not hold for constrained neural networks using monotonone convex non-saturated activation functions: the resulting model are convex and cannot approximate non-convex functions.

\section{Constrained neural networks}


\begin{figure}
    \begin{subfigure}[t]{0.32\textwidth}
        \includegraphics[width=\textwidth]{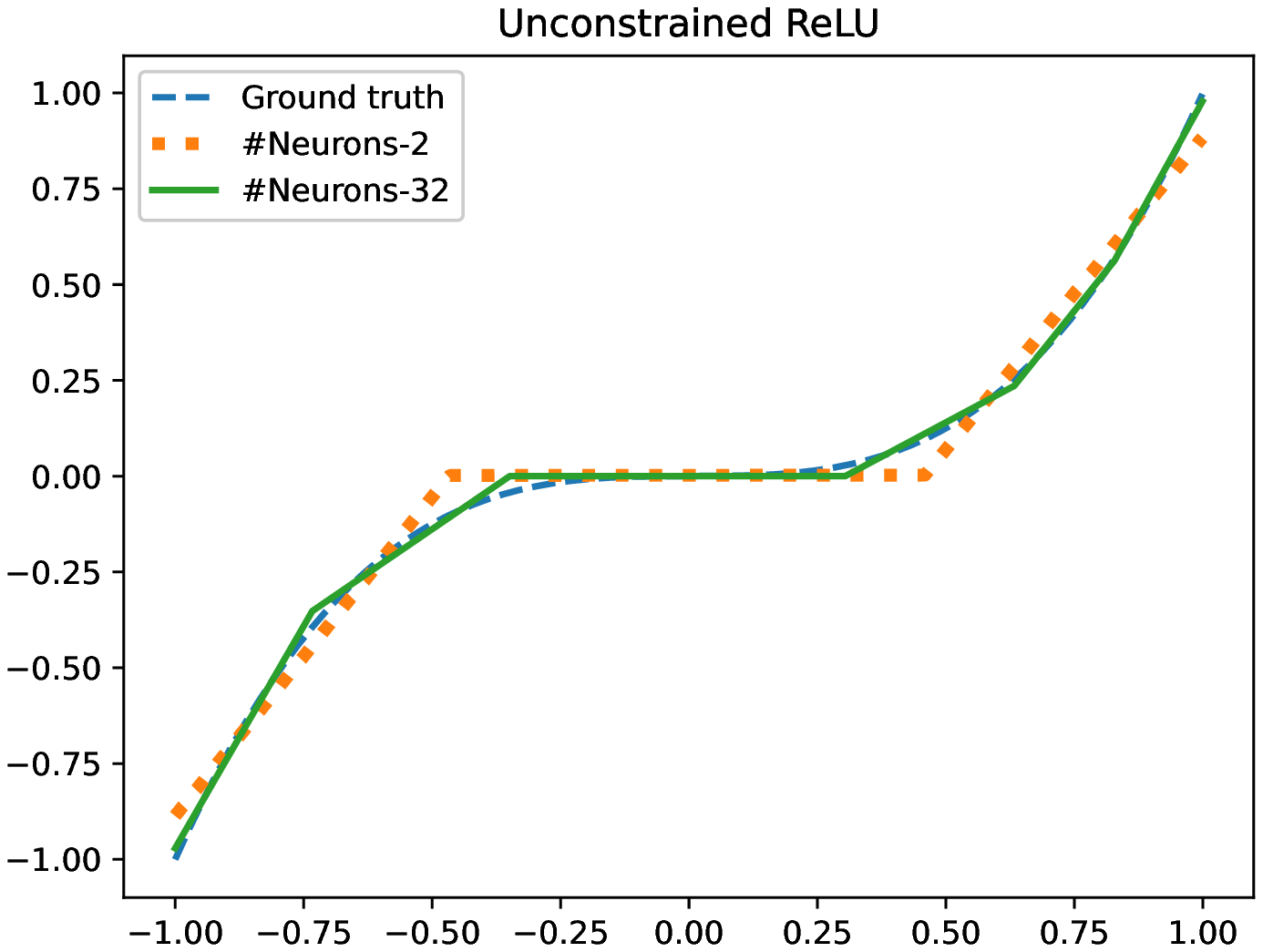}
        \caption{Unconstrained ReLU}
    \label{subfig:uncons_relu}
    \end{subfigure}
    \hfill
    \begin{subfigure}[t]{0.32\textwidth}
        \includegraphics[width=\textwidth]{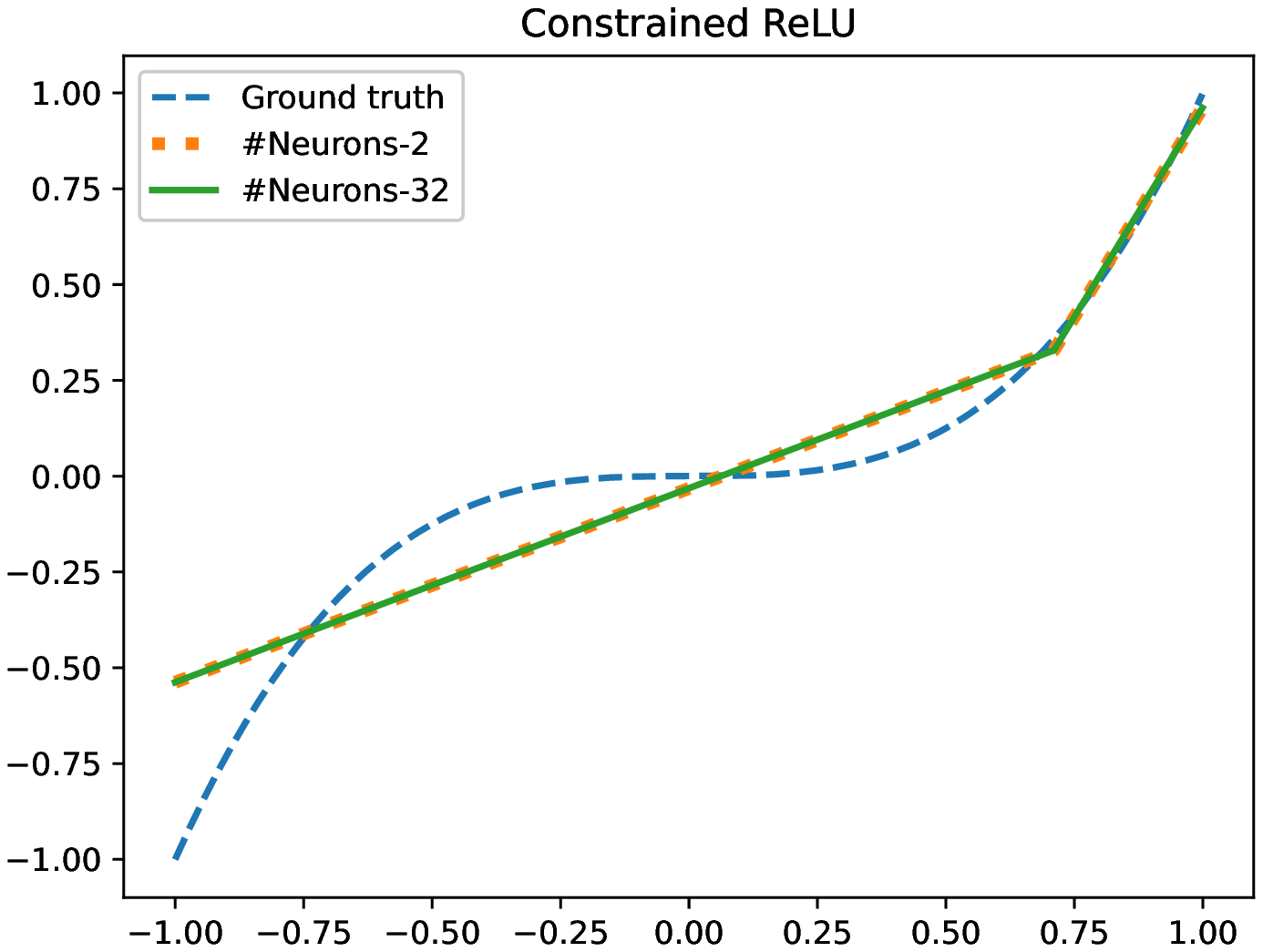}
        \caption{Constrained ReLU}
        \label{subfig:cons_relu}
    \end{subfigure}
    \hfill
    \begin{subfigure}[t]{0.32\textwidth}
        \includegraphics[width=\textwidth]{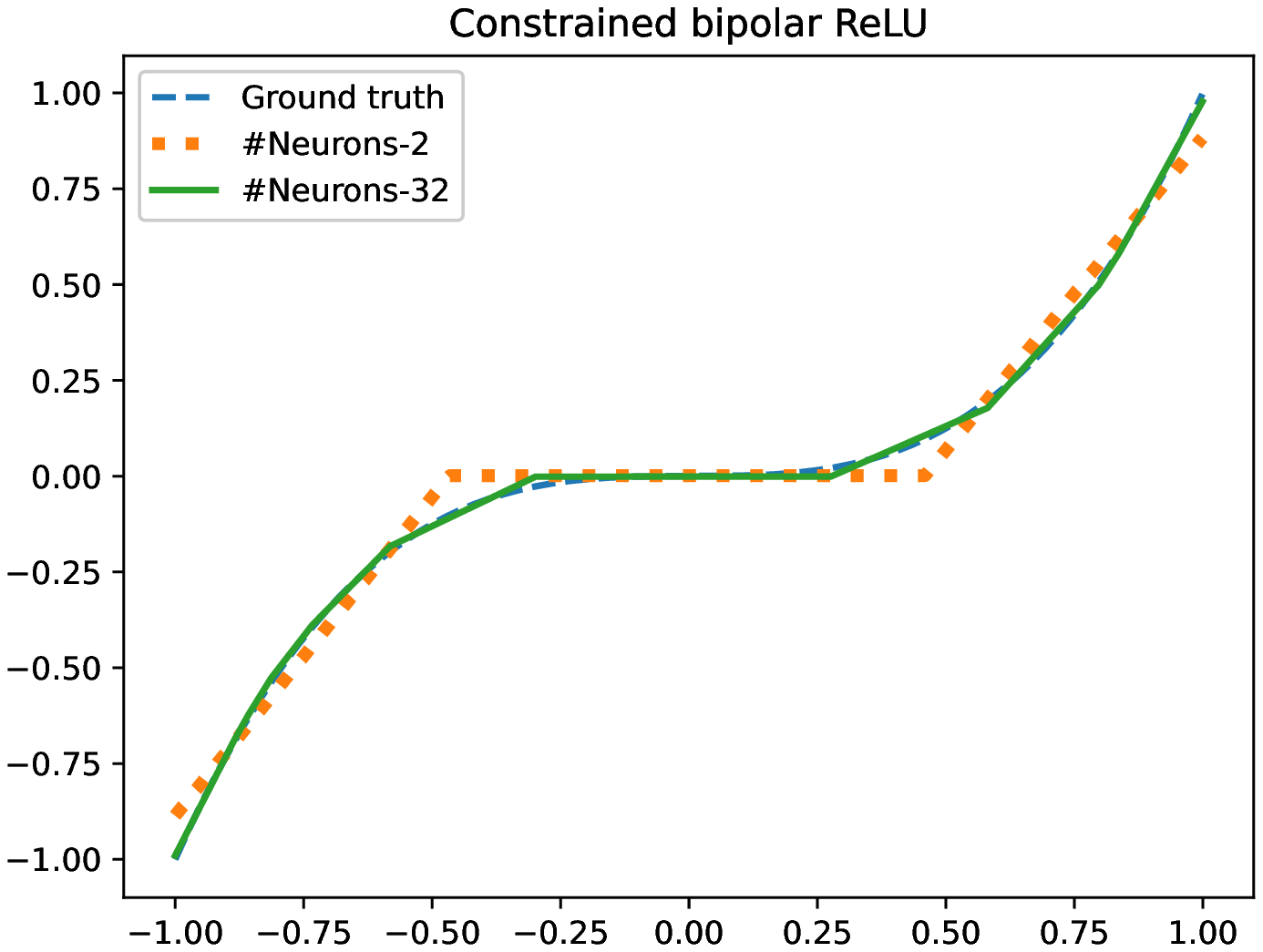}
        \caption{Constrained bipolar ReLU}
        \label{subfig:cons_birelu}
    \end{subfigure}
    \hfill
    
    \caption{Approximations of the cubic function $f(x)=x^3$}
    \label{fig:example_prob}
\end{figure}


%
Most of the commonly used activation functions such as ReLU, ELU, SELU etc. are  monotonically increasing, convex, non-polynomial functions. When used in a fully-connected, feed-forward neural network with at least one hidden layer and with unconstrained weights, they can approximate any continuous function on a compact subset. The simplest way to construct a monotonic neural network is to constrain its weights when used in conjunction with a monotone activation function. However, when the activation function is convex as well, the constrained neural network is not able to approximate non-convex functions. 

To better illustrate this, and to propose a simple solution in this particular example, we refer the readers to the Figure \ref{fig:example_prob} where the goal is to approximate a simple cubic function $x^3$ using a neural network with a single hidden layer with either $2$ or $32$ neurons and with ReLU activation. A cubic function is apt for our illustration since it is concave in the considered interval $[-1, 0]$ and convex in the interval $[0, -1]$:
\begin{itemize}
    \item [Fig. \ref{subfig:uncons_relu}.] An unconstrained ReLU network with $n$ neurons can approximate both concave and convex segments of the cubic function using at most $n+1$ piecewise linear segments. Increasing the number of neurons will provide a better fit with the function being approximated. Notice that even though the cubic function is monotone, there is no guarantee that the trained model will be monotone as well.
    
    \item [Fig. \ref{subfig:cons_relu}.] If we constrain the weights of the network to be non-negative while still employing ReLU activation, the resulting model is monotone and convex. We can no longer approximate non-convex segments such as the cubic function on $[-1, 0]$ in the figure and increasing the number of neurons from $2$ to $32$ does not yield any significant improvement in the approximation.
    
    \item[Fig. \ref{subfig:cons_birelu}.] The proposed solution uses both convex and concave activation functions in the hidden layer, in this case bipolar ReLU, to gain ability to model non-convex, monotone continuous functions. Notice that increasing the number of neurons increases the number of piecewise linear segments to approximate the cubic function. The resulting network is monotone by construction even when trained on noisy data.
\end{itemize}

\subsection{Constrained monotone fully connected layer}

Our construction is preconditioned on a priori knowledge of (partial) monotonicity of a multivariate, multidimensional function $f$. Let $f: K \mapsto \mathbb{R}^m$ be defined on a compact segment $K \subseteq \mathbb{R}^n$. Then we define its $n$-dimensional \emph{monotonicity indicator vector} $\mathbf{t}$ element wise as follows:
\begin{equation}
    t_i= \begin{cases}
      1 & \text{if }\cfrac{\partial f(\mathbf{x})_j} {\partial x_i} \geq 0 \medspace
    \text{for each } j \in \{1, \dots , m\}\\
      -1 & \text{if }\cfrac{\partial f(\mathbf{x})_j} {\partial x_i} \leq 0  \medspace
    \text{for each } j \in \{1, \dots , m\}\\
      0 & \text{otherwise}
    \end{cases} 
    \: 
\end{equation}

Given an $(n \times m)$-dimensional matrix $\mathbf{W}$ and $n$-dimensional monotonicity indicator vector $\mathbf{t}$, we define the operation $|.|_{t}$ assigning an $(n \times m)$-dimensional matrix $\mathbf{W'} = |\mathbf{W}|_{t}$ to $\mathbf{W}$ as follows:
\begin{equation}
\label{eq:enforce_sign}
    w'_{i,j}= \begin{cases}
      |w_{i,j}| & \text{if }t_i=1\\
      -|w_{i,j}| & \text{if }t_i=-1\\
      w_{i,j} & \text{otherwise}
    \end{cases} 
\end{equation}

\begin{definition}[Constrained linear layer]
Let $\mathbf{W}$ be an $(n \times m)$-dimensional matrix, $\mathbf{x}$ an $n$-dimensional vector, $\mathbf{b}$ an $m$-dimensional vector and $\mathbf{h}$ an $m$-dimensional vector. Then the output $\mathbf{h}$ of the \emph{constrained linear layer} with monotonicity indicator vector $\mathbf{t}$, weights $\mathbf{W}$, biases $\mathbf{b}$ and input $\mathbf{x}$ is:
\begin{equation}
\mathbf{h} = |\mathbf{W}|_{\mathbf{t}} \cdot \mathbf{x} + \mathbf{b}
\end{equation}
\end{definition}

\begin{lemma} \label{lem:mono_linear}
For each $i \in \{1, \dots, n\}$ and $j \in \{1, \dots, m\}$ we have:
\begin{itemize}
    \item if $t_i = 1$, then $\cfrac{\partial{h_j}}{\partial{x_i}} \geq 0$, and
    \item if $t_i = -1$, then $\cfrac{\partial{h_j}}{\partial{x_i}} \leq 0$.
\end{itemize}
\end{lemma}

Let $\breve{\rho}$ be a monotonically increasing \emph{convex} function such as ReLU, ELU, etc. Then its point reflection $\hat{\rho}$ around the origin is defined as:
\begin{equation}
    \hat{\rho}(x) = -\breve{\rho}(-x)
\end{equation}
Notice that $\hat{\rho}$ is a monotonically increasing \emph{concave} function.

\begin{definition}[Combined activation function]
Given a monotonically increasing \emph{convex} function $\breve{\rho}$, its point reflection $\hat{\rho}$ and $m$-dimensional \emph{activation selector vector} $\mathbf{s} \in [0, 1]^m$, the output of the \emph{combined activation function} $\rho^{\mathbf{s}}$ is a weighted sum of $\breve{\rho}$ and $\hat{\rho}$:
\begin{equation}
    \rho^{\mathbf{s}}(\mathbf{h}) = \mathbf{s} \odot \breve{\rho}(\mathbf{h}) + (\mathbf{1}-\mathbf{s}) \odot \hat{\rho}(\mathbf{h})
\end{equation}
where $\odot$ is the element-wise (Hadamard) product and $\mathbf{1}$ is $m$-dimensional vector with all elements equal to 1.

\end{definition}

\begin{lemma}  \label{lem:mono_activation}
Let $\mathbf{y} = \rho^{\mathbf{s}}(\mathbf{h})$. Then for each $j \in \{1, \dots, m\}$ we have $\cfrac{\partial{y_j}}{\partial{h_j}} \geq 0$. Moreover
\begin{itemize}
\item if $\mathbf{s} = \mathbf{1}$, then $\rho^{\mathbf{s}}_j$ is convex; and

\item if $\mathbf{s} = \mathbf{0}$, then $\rho^{\mathbf{s}}_j$ is concave.
\end{itemize}
\end{lemma}

\begin{definition}[Monotone constrained fully connected layer]
Let $n$ and $m$ be positive natural numbers, $\breve{\rho}$ a monotonically increasing \emph{convex} function, 
$\mathbf{t}$ an $n$-dimensional monotonicity selector vector and $\mathbf{s}$ an $m$-dimensional activation selector vector. Then the \emph{monotone constrained fully connected layer} with $m$ neurons is a tuple $(\breve{\rho}, \mathbf{t}, \mathbf{s})$, denoted $\mathcal{MFC}_{\breve{\rho}, \mathbf{s}, \mathbf{t}}$.

Moreover, let weights $\mathbf{W}$ be an $(n \times m)$-dimensional matrix, biases $\mathbf{b}$ an $m$-dimensional vector and input $x$ an $n$-dimensional vector. Then the \emph{output function} of the monotone constrained fully connected layer, denoted $\mathcal{MFC}_{\breve{\rho}, \mathbf{s}, \mathbf{t}}^{\mathbf{W}, \mathbf{b}}(\mathbf{x})$, assigns an $m$-dimensional vector $\mathbf{y}$ to $(\mathbf{W}, \mathbf{b}, \mathbf{x})$ as follows:
\begin{align}
    \mathbf{y} & = \mathcal{MFC}_{\breve{\rho}, \mathbf{s}, \mathbf{t}}^{\mathbf{W}, \mathbf{b}}(\mathbf{x}) = \rho^{\mathbf{s}}\left(|\mathbf{W}|_{\mathbf{t}} \cdot \mathbf{x} + \mathbf{b}\right)
\end{align}
\end{definition}

Notice that $\mathcal{MFC}_{\breve{\rho}, \mathbf{s}, \mathbf{t}}$ determines an architecture of the layer, while $\mathcal{MFC}_{\breve{\rho}, \mathbf{s}, \mathbf{t}}^{\mathbf{W}, \mathbf{b}}$ is a fully instantiated layer with all of its internal parameters $\mathbf{W}$ and $\mathbf{b}$ defining its output function.

From Lemma \ref{lem:mono_linear} and \ref{lem:mono_activation} directly follows:
\begin{corollary}
For each $i \in \{1, \dots, n\}$ and each $j \in \{1, \dots, m\}$ we have:
\begin{itemize}
    \item if $t_i = 1$, then $\cfrac{\partial{y_j}}{\partial{x_i}} \geq 0$;
    \item if $t_i = -1$, then $\cfrac{\partial{y_j}}{\partial{x_i}} \leq 0$;
    \item if $\mathbf{s} = \mathbf{1}$, then $\mathbf{y}_j$ is convex; and
    \item if $\mathbf{s} = \mathbf{0}$, then $\mathbf{y}_j$ is concave.
\end{itemize}
\end{corollary}

\begin{figure}[!ht]
    \centering
    \includegraphics[width=0.9 \textwidth]{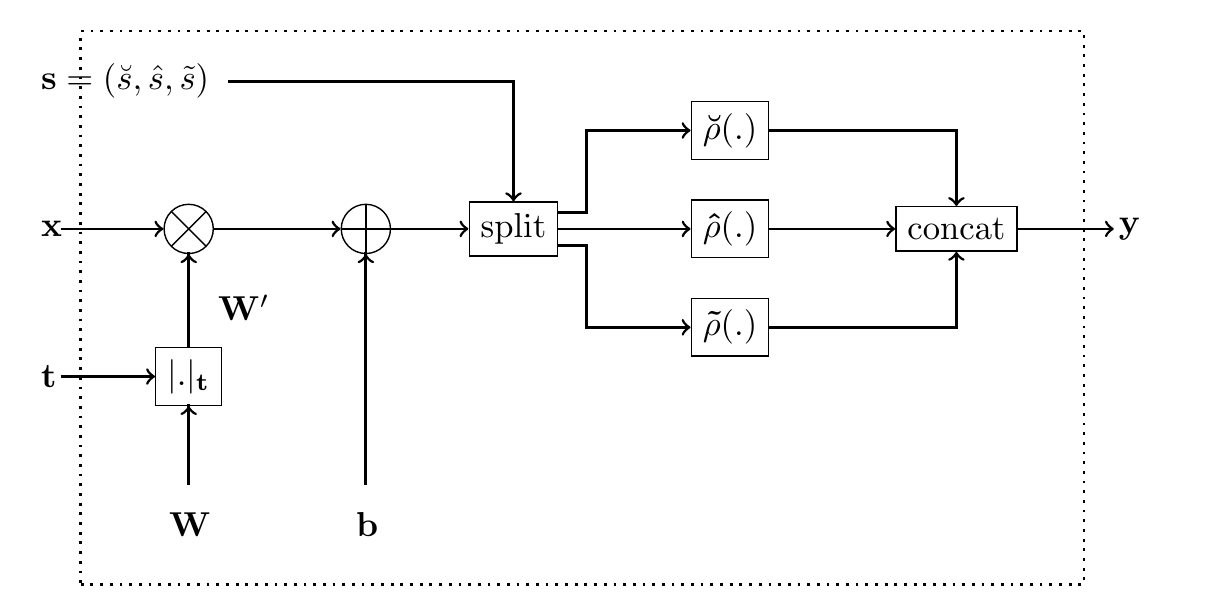}
    \caption{Proposed Monotonic Dense Unit}
    \label{fig:mono_dense}
\end{figure}

On the layer level, we can control both monotonicity, convexity and concavity of the output with respect to chosen input variables. The following section discuss how we can use such layers to build practical neural networks with the same properties.

\subsection{Building deep neural network by composition of monotonic dense units}

As mentioned before, the main advantage of our proposed monotonic dense unit is its simplicity. We can build deep neural nets with different architectures by plugging in our monotonic dense blocks. The figures Fig \ref{subfig:neural_type_1} and \ref{subfig:neural_type_2}  shows two examples of neural architectures that can be built using the proposed monotonic dense block. 

\begin{figure}
    \begin{subfigure}[t]{0.98\textwidth}
        \includegraphics[width=\textwidth]{Images/Composition of Monotonic Dense.pdf}
        \caption{Neural architecture type 1}
    \label{subfig:neural_type_1}
    \end{subfigure}
    \hfill
    \begin{subfigure}[t]{0.98\textwidth}
        \includegraphics[width=\textwidth]{Images/Composition of Monotonic Dense2.pdf}
        \caption{Neural architecture type 2}
        \label{subfig:neural_type_2}
    \end{subfigure}
    \hfill
    \caption{Examples of two types of neural architectures built using our proposed monotonic dense units}
    \label{fig:example_composition}
\end{figure}

The figure Fig. \ref{subfig:neural_type_1} shows an example of a model with both monotonic feature and non-monotonic feature in the input. For non-monotonic feature, we can employ any kind of neural network layer such as simple fully connected or dense layer, or embedding neural network layer or even complicated block such as transformer etc. For the sake of simplicity, we show neural architecture example with only dense layer. For monotonic feature, we specifically employ the proposed monotonic dense unit. It should be noted that our proposed monotonic dense unit would act as a regular dense layer without any constraint on weights if the indicator vector $\mathbf{t}$ is $0$. In this specific example, there is only one hidden monotonic dense layer and the bipolar activation is used in the hidden layer.  For the final output monotonic dense layer we do not employ any activation but only the weights are constrained to be non-negative to preserve monotonicity. Depending on whether the problem at hand is a regression problem or a classification problem or even a multi-task problem we employ appropriate final activation function (such as linear activation or sigmoid or softmax) to obtain the final output.

The figure Fig. \ref{subfig:neural_type_2} shows another example of a model with monotonically increasing and convex feature, monotonically decreasing and concave feature, and non-monotonic feature in the input. Each of the input feature has a separate dense layer associated with it. The weights of the monotonic dense unit associated with the first feature is constrained to be positive and a convex monotonic activation function is used as is. The weights of the monotonic dense unit associated with the first feature is constrained to be negative and point reflection (or mirror) of a convex monotonic activation function is used. The third feature has a dense layer with unconstrained weights associated with it since it is non-monotonic. The weights of the subsequent  neural layers are constrained to be non-negative just like the example above to enforce monotonicity in the whole network.

 \section{Experiments}\label{sec:experiments}
 
 
We verify our method in various practical
settings and datasets. 
Experiment results show that networks learned by our method can achieve higher test accuracy with fewer parameters, than  
the best-known algorithms for monotonic neural networks, including Min-Max Network~\citep{daniels2010monotone} and Deep Lattice Network~\citep{you2017deep}. 
Our method also outperforms 
traditional monotonic methods, such as isotonic regression and monotonic XGBoost, in accuracy.  

\paragraph{Datasets:} Experiments are performed on three datasets: COMPAS~\citep{compas}, Blog Feedback Regression~\citep{buza2014feedback} and Loan Defaulter\footnote{https://www.kaggle.com/wendykan/lending-club-loan-data}. 
\emph{COMPAS} is a classification dataset with 13 features, 4 of them being monotonic. \emph{Blog Feedback} is a regression dataset with 276 features, 8 of them being monotonic. \emph{Loan Defaulter} is a classification dataset with 28 features, 5 of them being monotonic. 
 
\paragraph{Methods for Comparison:} We compare our method with six methods that can generate partially monotonic models. \emph{Isotonic Regression}: a deterministic method for monotonic regression~\citep{dykstra2012advances}. \emph{XGBoost}: a popular algorithm based on gradient boosting decision tree~\citep{xgboost}. \emph{Crystal}: an algorithm using ensemble of lattices~\citep{milani2016fast}. \emph{Deep Lattice Network (DLN)}: a deep network with ensemble of lattices layer~\citep{you2017deep}. 
\emph{Non-Neg-DNN}: deep neural networks with non-negative weights. 
\emph{Min-Max}: a classical three-layer network with one linear layer, one min-pooling layer, and one max-pooling layer~\citep{daniels2010monotone}.
 
 \paragraph{Hyperparameter Validation:} We use the datasets employed by authors in \cite{liu2020certified} and use the same train and test split for fair comparison. For the classification tasks we use cross-entropy and for regression tasks we use mean-squared-error as loss functions. We employ grid search to find the optimal number of neurons, network depth or layers, batch size, activation function and the number of epochs. For training, we first find the optimal learning rate using learning rate finder \cite{smith2017cyclical} and then train with one cycle policy \cite{smith2018disciplined}.

 
The results on the dataset
above are summarized in Tables 1, 2, 3. It shows that our method tends either match or surpass
the other methods in terms of test accuracy for classification tasks and root mean squared error (RMSE) for regression tasks.

\begin{table}[htbp]
\centering
\renewcommand\arraystretch{1.3}
\small{
\begin{tabular}{l||ll|ll|ll}
&Data:& COMPAS &Data:& Blog feedback &Data:& Loan defaulter
\\
 Method & Params & Test Acc $\uparrow$ & Params & RMSE $\downarrow$ & Params & Test Acc $\uparrow$ \\\hline
 Isotonic& N.A. & 67.6\% & N.A. & 0.203  & N.A. & 62.1\% \\
 XGBoost~\citep{xgboost} & N.A. & 68.5\% $\pm$ 0.1\%  & N.A. & 0.176 $\pm$ 0.005  & N.A. & 63.7\% $\pm$ 0.1\%\\
 Crystal~\citep{milani2016fast} & 25840 & 66.3\% $\pm$ 0.1\%  & 15840 & 0.164 $\pm$ 0.002 & 16940 & 65.0\% $\pm$ 0.1\% \\
 DLN~\citep{you2017deep}     & 31403 & 67.9\% $\pm$ 0.3\% & 27903 & 0.161 $\pm$ 0.001& 29949  & 65.1\% $\pm$ 0.2\% \\
 Min-Max~\citep{daniels2010monotone}     & 42000  & 67.8\% $\pm$ 0.1\% & 27700 & 0.163 $\pm$ 0.001  & 29000 & 64.9\% $\pm$ 0.1\%\\
  Non-Neg-DNN & 23112 & 67.3\% $\pm$ 0.9\% & 8492  & 0.168 $\pm$ 0.001 & 8502 & 65.1\% $\pm$ 0.1\%  \\
 Cert-Mono~\citep{liu2020certified} & 23112 & 68.8\% $\pm$ 0.2\%  & 8492 & 0.158 $\pm$ 0.001   & 8502 &  65.2\% $\pm$ 0.1\%  \\
 \textbf{Ours}    & \textbf{101} & \textbf{68.9\% $\pm$ 0.5\%}& \textbf{1101} & \textbf{0.156 $\pm$ 0.001}   & \textbf{177} & \textbf{65.3\% $\pm$ 0.0003\%} \\ 
\end{tabular}}
\vspace{5pt}
\caption{Results on test datasets: COMPAS~\citep{compas}, Blog Feedback Regression~\citep{buza2014feedback} and Loan Defaulter}
\label{tab:compas}
\end{table}










 
\bibliographystyle{plain}

\bibliography{references}

\end{document}


\maketitle

\section{Detailed proofs}

We restate all lemmas from the main text here are give detailed proofs of them.

The following is well known result, proved here for completeness only:
\begin{lemma}
For each $i \in \{1, \dots, n\}$ and $j \in \{1, \dots, m\}$ we have:
\begin{itemize}
    \item if $t_i = 1$, then $\cfrac{\partial{h_j}}{\partial{x_i}} \geq 0$, and
    \item if $t_i = -1$, then $\cfrac{\partial{h_j}}{\partial{x_i}} \leq 0$.
\end{itemize}
\end{lemma}
\begin{proof}
From equation \ref{eqn:constraned_lin_layer}, we have
\begin{align}
\mathbf{h} & = |\mathbf{W}^T|_{\mathbf{t}} \cdot \mathbf{x} + \mathbf{b} \\
h_j & = \sum_{i} w'_{i,j} x_i + b_j \\
\cfrac{\partial{h_j}}{\partial{x_i}} & = w'_{i, j}
\end{align}

Finally, from equation \ref{eq:enforce_sign} we have 
\begin{equation*}
\cfrac{\partial{h_j}}{\partial{x_i}} =
    \begin{cases}
      |w_{i,j}| \geq 0 & \text{if }t_i=1\\
      -|w_{i,j}| \leq 0 & \text{if }t_i=-1
    \end{cases} 
\end{equation*}
\end{proof}

\begin{lemma} 
Let $\mathbf{y} = \rho^{\mathbf{s}}(\mathbf{h})$. Then for each $j \in \{1, \dots, m\}$ we have $\cfrac{\partial{y_j}}{\partial{h_j}} \geq 0$. Moreover
\begin{itemize}
\item if $\mathbf{s} = (m, 0, 0)$, then $\rho^{\mathbf{s}}_j$ is convex; and

\item if $\mathbf{s} = (0, m, 0)$, then $\rho^{\mathbf{s}}_j$ is concave.
\end{itemize}
\end{lemma}

\begin{proof}
From equation \ref{eq:activation_concave},\ref{eq:activation_saturated} and \ref{eqn:combined_activaion}:
\begin{align*}
    \hat{\rho}(x) & = -\breve{\rho}(-x) \\
    \tilde{\rho}(x) & = \begin{cases}
      \breve{\rho}(x+1)-\breve{\rho}(1) & \text{if }x < 0\\
      \hat{\rho}(x-1)-\hat{\rho}(1) & \text{otherwise}
    \end{cases} 
\end{align*}
\begin{equation*}
    \rho^{\mathbf{s}}(h_j) = \begin{cases}
        \breve{\rho}(h_j) & \text{if } j \leq \breve{s} \\
        \hat{\rho}(h_j) & \text{if } \breve{s} < j \leq \hat{s} \\
        \tilde{\rho}(h_j) & \text{otherwise} 
    \end{cases}
\end{equation*}
we have:
\begin{equation*}
    \cfrac{\partial{y_j}}{\partial{h_j}} = \begin{cases}
        {\breve\rho}'(h_j) \geq 0 & \text{if } j \leq \breve{s} \\
        {\breve\rho}'(-h_j)  \geq 0 & \text{if } \breve{s} < j \leq \hat{s} \\
        {\breve\rho}'(h_j+1) \geq 0 & \text{if } \breve{s} + \hat{s} < j \text{ and } h_j < 0 \\
        {\breve\rho}'(1-h_j) \geq 0 & \text{if } \breve{s} + \hat{s} < j \text{ and } h_j \geq 0 \\
    \end{cases}
\end{equation*}

if $\mathbf{s} = (m, 0, 0)$, we have:
$$
\rho^{\mathbf{s}}(h_j) = \breve{\rho}(h_j) 
$$
which is a convex function. 

Similarly, if $\mathbf{s} = (0, m, 0)$, we have:
$$
\rho^{\mathbf{s}}(h_j) = \hat{\rho}(h_j) 
$$
which is a concave function.
\end{proof}

\begin{corollary}
For each $i \in \{1, \dots, n\}$, $j \in \{1, \dots, m\}$ we have:
\begin{itemize}
    \item if $t_i = 1$, then $\cfrac{\partial{y_j}}{\partial{x_i}} \geq 0$,
    
    \item if $t_i = -1$, then $\cfrac{\partial{y_j}}{\partial{x_i}} \leq 0$, 
    
    \item if $\mathbf{s} = (m, 0, 0)$, then $\mathbf{y}_j$ is convex; and

    \item if $\mathbf{s} = (0, m, 0)$, then $\mathbf{y}_j$ is concave.
    
\end{itemize}
\end{corollary}

For completeness, we repeat the Theorem 3.1 from \cite{daniels2010monotone} and its proof here:

\begin{theorem}
For any continuous monotone nondecreasing function $f: K \xrightarrow{} \mathbb{R}$, where $K$  is a compact subset of $\mathbb{R}^k$, there exists a feedforward neural network using the sigmoid as the activation function with at most $k$ hidden layers, positive weights, and output $O$  such that $|O_x - f(x) | < \epsilon$, for any $x \in K$ and $\epsilon > 0$. 
\end{theorem}

\begin{proof}
The proof is derived by induction on the number of input variables $k$. Without loss of generality, we may assume that $f > 0$ (otherwise, we add a constant $C$  and approximate $f+C$
with the network output $O$, then modify $O$  with a negative bias at the output node). First, we assume that $f$ is strictly increasing and $C^{\infty}$. In case of $k=1$, we write
\begin{equation}
\label{eqn:heavyside_approx}
f(x) = \int_0^{\infty} \mathbf{H}\left(f(x) - u\right) du
\end{equation}
where $\mathbf{H}$ is the Heavyside function:
$$
H(x) = \left\{ \begin{array}{cc}
1     &  \textnormal{if } x\geq 0 \\
0     & \textnormal{otherwise}
\end{array} \right.
$$

Since $f$ is continuous and increasing, it is invertible and therefore the right-hand side of \ref{eqn:heavyside_approx} can be written as
\begin{equation}
\label{eqn:heavyside_inverted}
f(x) = \int_0^{\infty} \mathbf{H}\left(x - f^{-1}(v)\right) dv
\end{equation}
The integral can be approximated arbitrarily well by a Riemann sum
\begin{equation}
\label{eqn:riemann}
\sum_{i=1}^{N}(v_{i+1}-v{i})\ \mathbf{H}\left(x - f^{-1}(v_i)\right)
\end{equation}
where $[v_i]_{i=1}^N$ is a partition of the interval $[f(a), f(b)]$. This  expression corresponds to a neural network with input $x$, one hidden layer with $N$ neurons all connected to the input with weight of 1, bias term in the hidden neurons $f^{-1}(v_i)$, and the weights connecting the hidden layer with the output $v_{i+1} - v{i} > 0$. Note that the Heavyside function $\mathbf{H}$ can be replaced by a sigmoid activation function using a standard approximation argument.

Assume that Theorem 3.1 holds for $k-1$ input variables. We now combine the integral representation in \ref{eqn:heavyside_approx} with the induction assumption. For a given $v$, we may solve the equation of the level set corresponding to to $v$ for $x_k$
$$
f(x_1, \dots, x_k) = v.
$$
By the implicit function theorem, there exists a function $g_v$ such that
\begin{equation}
\label{eqn:implicit_function_thm}
f(x_1, \dots, g_v(x_1, \dots, x_{k-1})) = v.
\end{equation}
Note that $g_v$ is decreasing in all arguments $x_i$. This can be seen by taking the partial derivative of \ref{eqn:implicit_function_thm} with respect to $x_i$. We will now show that
$$
\mathbf{H}\left(f(x) - v \right) = \mathbf{H}\left(x_k - g_v(x_1, \dots, x_{k-1}) \right)
$$
analogously to \ref{eqn:heavyside_inverted} for the 1-D case. Note that it is sufficient to show that
$$
f(x) < v \textnormal{ if and only if } x_k < g_v(x_1, \dots, x_{k-1})
$$
and 
$$
f(x) > v \textnormal{ if and only if } x_k > g_v(x_1, \dots, x_{k-1}) .
$$
But this follows easily from \ref{eqn:implicit_function_thm} and the fact that $f$ is increasing in all its arguments. We now approximate the integral in \ref{eqn:heavyside_approx} with a Riemann sum, leading to the following equation analogously to \ref{eqn:riemann}:
\begin{equation}
\label{eqn:riemann_inductive}
    R = \sum_{i=1}^N(v_{i-1} - v_{i})\ \mathbf{H}\left(x_k - g_{v_i}\left(x_1, \dots, x_{k-1} \right)\right)
\end{equation}
Since $g_{v_i}$ is decreasing in all its arguments $-g_{v_i}$ is increasing. By the induction assumption, we can approximate   $-g_{v_i}$  with a feedforward neural network $O_i$ with $x_1, \dots, x_{k-1}$ as inputs, $k-1$ hidden layers, and nonnegative weights, such that
$$
\left| \sum_{i=1}^N(v_{i-1} - v_{i})\ \mathbf{H}\left(x_k - O_i\left(x_1, \dots, x_{k-1} \right)\right)  - R \right| < \epsilon
$$
because the sum is finite. Expression \ref{eqn:riemann_inductive}  corresponds to a feedforward neural network with $k$ inputs and $k$ hidden layers. Here      $k-1$ hidden layers are needed to represent $-g_{v_i}$
and the  $k$-th hidden layer is needed to combine $N$ neural networks with outputs $O_i$ and the input $x_k$. The weights on the connections between the last hidden layer and the final output are $(v_{i+1} - v_i) > 0$. The input $x_k$ is directly (skip-layer) connected to the   $k$-th hidden layer.

We can now easily generalize the proof to continuous nondecreasing functions. For continuous functions, we define the convolution of $f$ with a mollifier $K_{\delta}$ by
$$
f_{\delta} = f \otimes K_{\delta}\
$$
Then, $f_{\delta}$ is $C^{\infty}$ and $f_\delta \longrightarrow f$ as $\delta \downarrow 0$ uniform on compact subsets. Furthermore, $f_{\delta}$ is also increasing since $K_{\delta} > 0$. Now choose $\delta$ such that $|f - f_\delta | < \frac{\epsilon}{2}$ and approximate  $f_{\delta}$   with a feedforward neural network $O$ such that $|f_\delta - O | < \frac{\epsilon}{2}$. Then, $|f - O | < \epsilon$.

If $f$ is nondecreasing, then approximate $f$ by $f_\delta$
$$
f_\delta = f + \delta(x_1 + \dots + x_k)
$$
which is strictly increasing and let $\delta \downarrow 0$.
 
\end{proof}

\begin{lemma}
Let $\breve{\rho} \in \breve{\mathcal{A}}$. Then the Heavyside function can be approximated with $\tilde{\rho}_H $ on $\mathbb{R}$, where
$$
\tilde{\rho}_H(x) = \alpha \tilde\rho(x) + \beta
$$
for some $\alpha, \beta \in \mathbb{R}$ and $\alpha > 0$.
\end{lemma}

\begin{proof}
    Since $\breve{\rho}$ is bounded from below, then it has a limit
    $$
    c = \lim_{x \to -\infty} \breve{\rho}(x) = -{\lim_{x \to \infty}} \hat{\rho}(x) < 0
    $$
    From equation \ref{eq:activation_saturated}, we have
    \begin{align*}
    {\lim_{x \to -\infty}} \tilde{\rho}(x) & = {\lim_{x \to -\infty}} \breve{\rho}(x) - \breve{\rho}(1) = c -\breve{\rho}(1)\\
    {\lim_{x \to +\infty}} \tilde{\rho}(x) & = {\lim_{x \to \infty}} \hat{\rho}(x) + \breve{\rho}(1) = - (c - \breve{\rho}(1))
    \end{align*}

    Let $\tilde{\rho}_H$ be defined as follows:
    $$
    \tilde{\rho}_H(x) = \cfrac{\tilde{\rho}(x) - c + \breve{\rho}(1)}{2 \left( -c + \breve{\rho}(1) \right)}
    $$
    Then 
    $$
    \lim_{a \to \infty}{\tilde{\rho}_H(a \cdot x)} = H(x)
    $$
\end{proof}

\begin{lemma}
    Let $\tilde{\rho}_{\alpha, \beta}$ be an activation function for some $\alpha, \beta \in \mathbb{R}, \alpha>0$ such that for every $x \in \mathbb{R}$
    $$
        \tilde{\rho}_{\alpha, \beta}(x) = \alpha \tilde{\rho}(x) + \beta .
    $$
    Then for every constrained monotone neural network $\mathcal{N}_{\alpha, \beta}$ using $\tilde{\rho}_{\alpha, \beta}$ as an activation function ($s=(0, 0, \tilde{s})$), there is a constrained monotone neural network $\mathcal{N}$ using $\tilde{\rho}$ as an activation function such that for every $\mathbf{x} \in \mathbb{R}^{n}$:
    $$
        \mathcal{N}(\mathbf{x}) = \mathcal{N}_{\alpha, \beta}(\mathbf{x}) .
    $$
    
\end{lemma}
\begin{proof}
Let $\mathbf{h}_k$ be the output of the $k$-th constrained linear layer of $\mathcal{N}_{\alpha, \beta}$ evaluated on input $\mathbf{x}$. For every $k>1$, we have
\begin{align*}
\mathbf{h}_1 & = |\mathbf{W}_1|_{\mathbf{t}_1} \cdot \mathbf{x} + \mathbf{b}_1 \\
\mathbf{h}_k & = |\mathbf{W}_k|_{\mathbf{t}_k} \cdot \mathbf{y}_{k-1} + \mathbf{b}_k \\
\mathbf{y}_k & = \tilde{\rho}_{\alpha, \beta}(\mathbf{h}_k) \\
\mathbf{y} & = \mathbf{h}_l
\end{align*}
where $l$ is the total number of layers of $\mathcal{N}_{\alpha, \beta}$.
We can rewrite $\mathbf{h}_k$ as follows:
\begin{align*}
\mathbf{h}_k & = |\mathbf{W}_k|_{\mathbf{t}_k} \cdot \mathbf{y}_{k-1} + \mathbf{b}_k \\
& = |\mathbf{W}_k|_{\mathbf{t}_k} \cdot \tilde{\rho}_{\alpha, \beta}(\mathbf{h}_{k-1}) + \mathbf{b}_k \\
& = |\mathbf{W}_k|_{\mathbf{t}_k} \cdot \left(\alpha \tilde{\rho}(\mathbf{h}_{k-1}) + \beta \mathbf{1} \right) + \mathbf{b}_k  & (\textnormal{with } |\mathbf{1}| = |\mathbf{h}_{k-1}|)\\
& = \alpha |\mathbf{W}_k|_{\mathbf{t}_k} \cdot  \tilde{\rho}(\mathbf{h}_{k-1}) + \beta |\mathbf{W}_k|_{\mathbf{t}_k} \mathbf{1} + \mathbf{b}_k \\
& = | \alpha \mathbf{W}_k|_{\mathbf{t}_k} \cdot  \tilde{\rho}(\mathbf{h}_{k-1}) + \beta |\mathbf{W}_k|_{t_k} \mathbf{1} + \mathbf{b}_k & (\textnormal{from } \alpha > 0)\\
& = | \mathbf{W}'_k|_{t_k} \cdot  \tilde{\rho}(\mathbf{h}_{k-1}) + \mathbf{b}'_k \\
\end{align*}
for $\mathbf{W}'_k = \alpha \mathbf{W}_k$ and $\mathbf{b}'_k = \beta |\mathbf{W}_k|_{t_k} \mathbf{1} + \mathbf{b}_k$.

Hence, for every $\mathbf{x} \in \mathbb{R}^{n}$, the output of the neural network $\mathcal{N}$ with weights $\mathbf{W}_1, \mathbf{W}'_2, \dots, \mathbf{W}'_l$ and biases $\mathbf{b}_1, \mathbf{b}'_2, \dots, \mathbf{b}'_l$ is:
$$
    \mathcal{N}(\mathbf{x}) = \mathcal{N}_{\alpha, \beta}(\mathbf{x})
$$

\end{proof}

\begin{theorem}
Let $\breve{\rho} \in \breve{\mathcal{A}}$. Then any multivariate continuous monotone function $f$ on a compact subset of $\mathbb{R}^k$ can be approximated with a monotone constrained neural network of at most $k$ layers using $\rho$ as the activation function.
\end{theorem}

\begin{proof}
The proof of the Theorem \ref{thm:universal_sigmoid} uses only the fact that the Heavyside function $\mathbf{H}$ defined as
$$
H(x) = \left\{ \begin{array}{cc}
1     &  \textnormal{if } x\geq 0 \\
0     & \textnormal{otherwise}
\end{array} \right.
$$
can be approximated with the sigmoid function on a closed interval (since $\lim\limits_{a \to \infty} \sigma(a x) = H(x)$).

From the Theorem \ref{thm:universal_sigmoid} and the Lemma \ref{lemma:heavyside_approx}, any continuous monotone function $f$ on a compact subset of $\mathbb{R}^k$ can be approximated with a monotone constrained neural network with at most $k$ layers using
$\tilde{\rho}_H$ as the activation function.

From Lemma \ref{lemma:lin_comb_activation}, we can replace $\tilde{\rho}_H$ with $\tilde{\rho}$. Hence, any continuous monotone function $f$ on a compact subset of $\mathbb{R}^k$ can be approximated with a monotone constrained neural network with at most $k$ layers using
$\rho$ as the activation function and $\mathbf{s}_i = (0, 0, \tilde{s}_i)$.



\end{proof}



\section{Datasets Description}

The descriptions of datasets used for comparison are detailed below. As mentioned in the section \ref{sec:experiments}, the datasets are chosen from \cite{liu2020certified} and \cite{sivaraman2020counterexample} for proper evaluation. The train-test splits of $80\%-20\%$ are used for all comparison experiments.

\begin{enumerate}
    \item COMPAS~\cite{compas} is a binary classification dataset, where the task is to predict risk score of an individual committing crime again two years, based on the criminal records of individuals arrested in Florida. The risk  score needs to be monotonically increasing with respect to the following attributes ~\texttt{number of prior adult convictions, number of juvenile felony, number of juvenile misdemeanor}, and \texttt{number of other convictions}. It should be noted that there have been ethical concerns with the dataset \cite{compas, Rudin2020Age}
    
    \item Blog Feedback~\citep{buza2014feedback} is a regression dataset where the task is to predict the number of comments in the upcoming 24 hours from a feature set containing 276 features of which 8 (A51, A52, A53, A54, A56, A57, A58, A59) are monotonic features. The readers are suggested to refer to link~\footnote{https://archive.ics.uci.edu/ml/datasets/BlogFeedback} for more details. As mentioned by the authors of \cite{liu2020certified}, only the data points with targets smaller than the 90th percentile are used since the outliers could dominate the mean-squared-error metric.
    
    \item Lending club loan data\footnote{https://www.kaggle.com/wendykan/lending-club-loan-data} is a classification dataset, where the task is to predict whether the individual would default on loan, from a feature set having 28 features containing data such as the current loan status, latest payment information etc,. The probability of default should be non-decreasing with respect to \texttt{number of public record bankruptcies, Debt-to-Income ratio}, and non-increasing with respect to \texttt{credit score, length of employment, annual income}. 
    
    \item Auto MPG\footnote{https://archive.ics.uci.edu/ml/datasets/auto+mpg} \cite{quinlan1993combining} is a regression dataset where the task is to predict  city-cycle fuel consumption in miles per gallon (MPG) from a feature set containing 7 features of which the monotonic features are \texttt{weight (W)}, \texttt{displacement (D)}, and \texttt{horse-power (HP)}

    \item Heart Disease\footnote{https://archive.ics.uci.edu/ml/datasets/heart+disease}   \cite{gennari1989models} is a classification dataset, where the task is to predict the presence of heart disease from a feature set containing 13 features of which the risk associated with heart disease needs to be monotonically increasing with respect to the features \texttt{trestbps (T), cholestrol (C))}
\end{enumerate}

\section{Additional Experiments and Results Details}

For all our experiments, we adopt simple architectures of the types depicted in Figure. \ref{subfig:neural_type_1} or Figure. \ref{subfig:neural_type_2}. More often than not, we have seen from our experiments that the neural architecture type 2 (Figure. \ref{subfig:neural_type_2}) performs better than the neural architecture type 1 (Figure. \ref{subfig:neural_type_1}). The number of hidden layers (apart from the input layer) is either set to be $1$ or $2$. The number of neurons in the hidden layer is selected from {4, 8, 16, 32, 64}. The activation function is set to be either exponential linear unit (ELU) or Rectified Linear Unit (ReLU).

\begin{figure*}[!tp]
    \begin{subfigure}[t]{0.24\textwidth}
        \includegraphics[width=\textwidth]{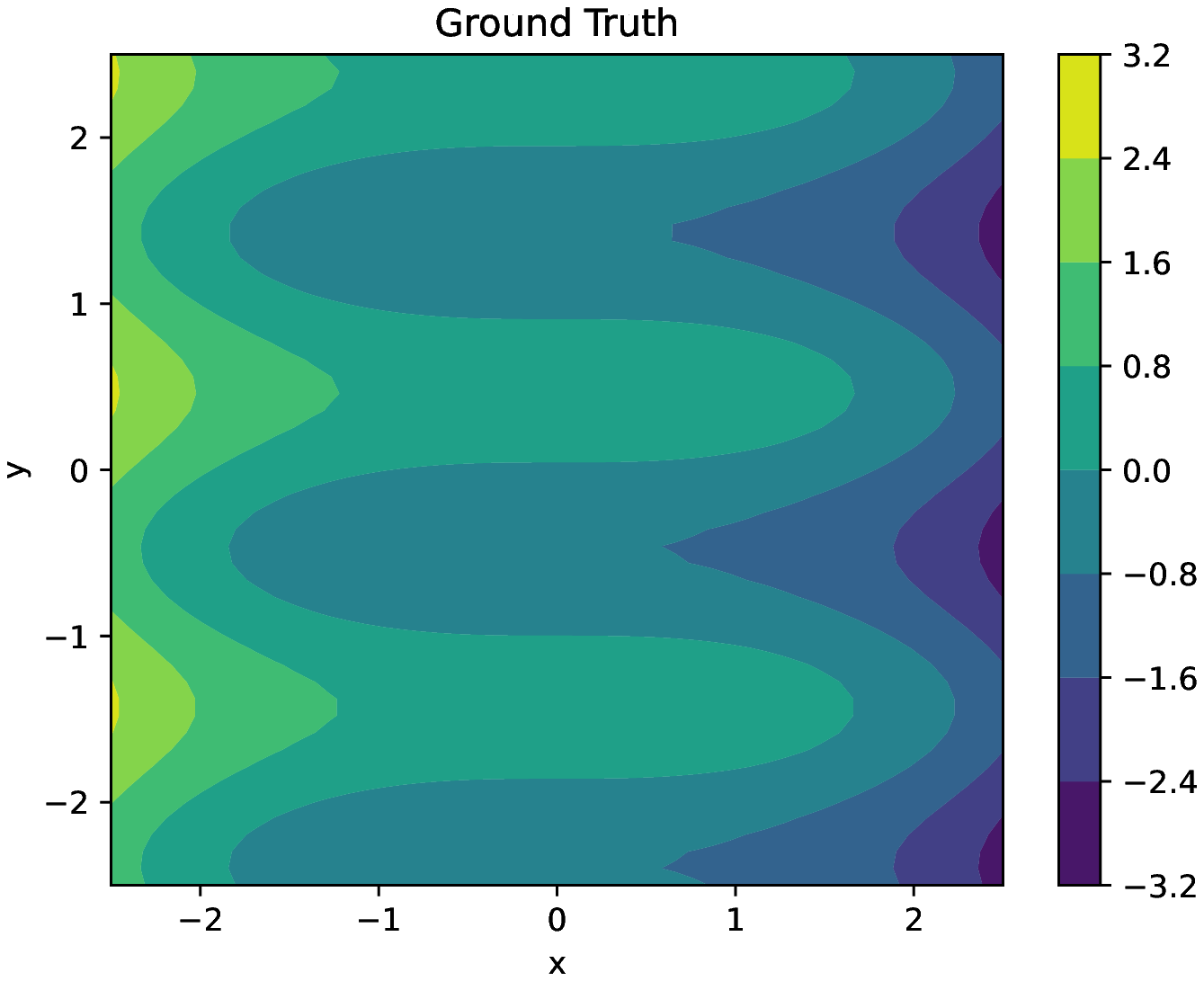}
        \caption{}
        \label{subfig:results_gt}
    \end{subfigure}
    \hfill
    \begin{subfigure}[t]{0.24\textwidth}
        \includegraphics[width=\textwidth]{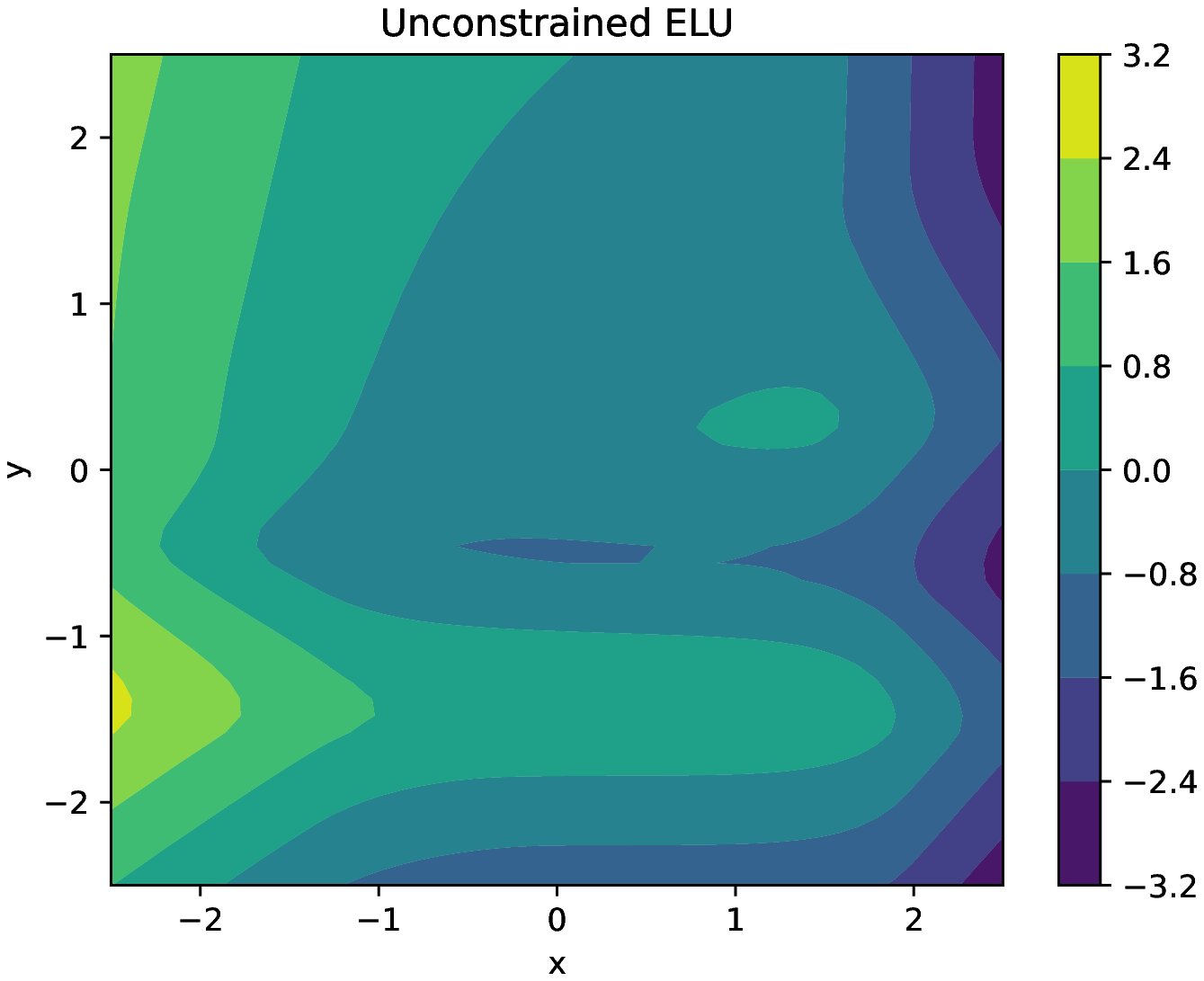}
        \caption{}
    \label{subfig:results_Unconstrained}
    \end{subfigure}
    \hfill
    \begin{subfigure}[t]{0.24\textwidth}
        \includegraphics[width=\textwidth]{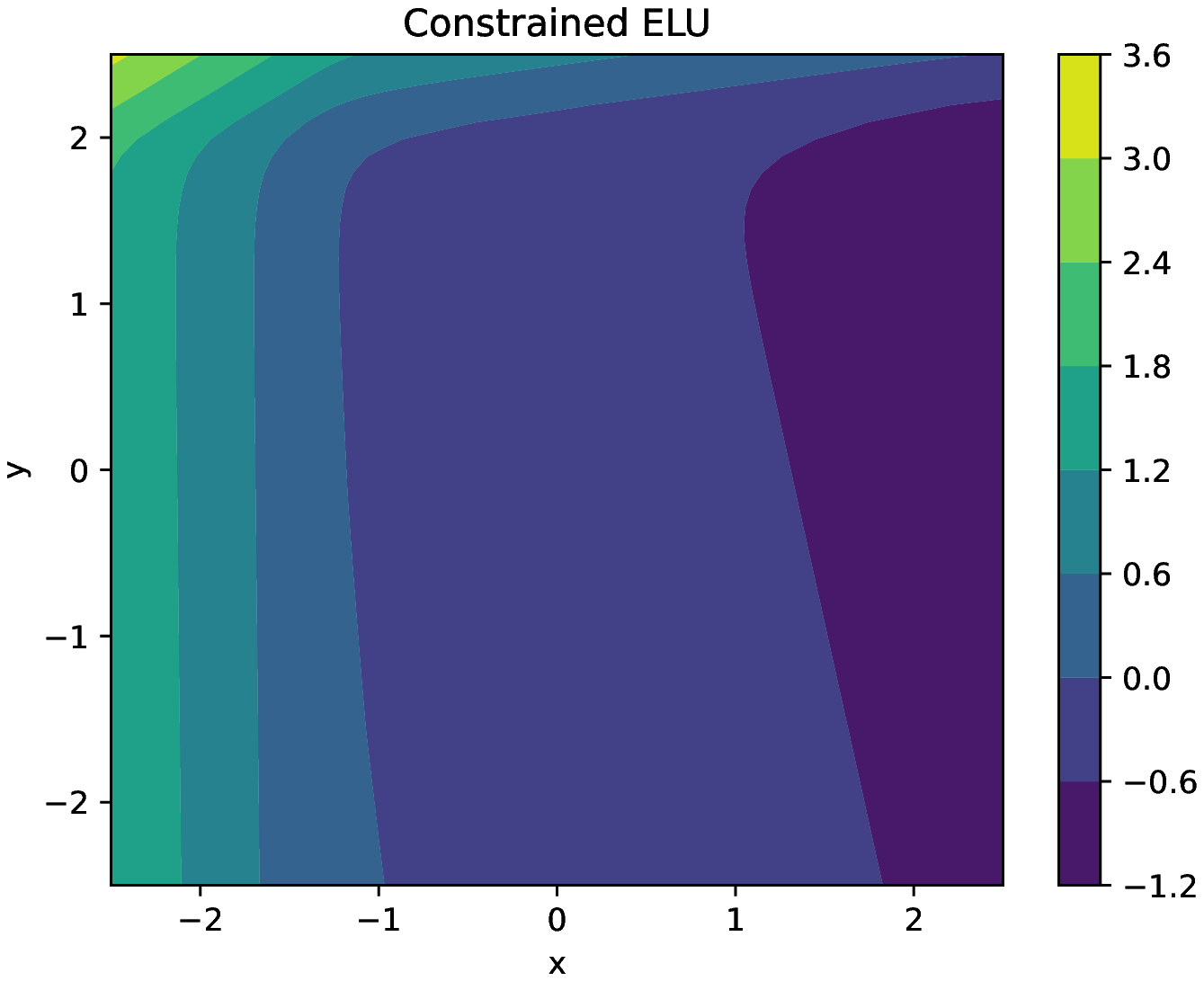}
        \caption{}
    \label{subfig:results_constrained}
    \end{subfigure}
    \hfill
    \begin{subfigure}[t]{0.24\textwidth}
        \includegraphics[width=\textwidth]{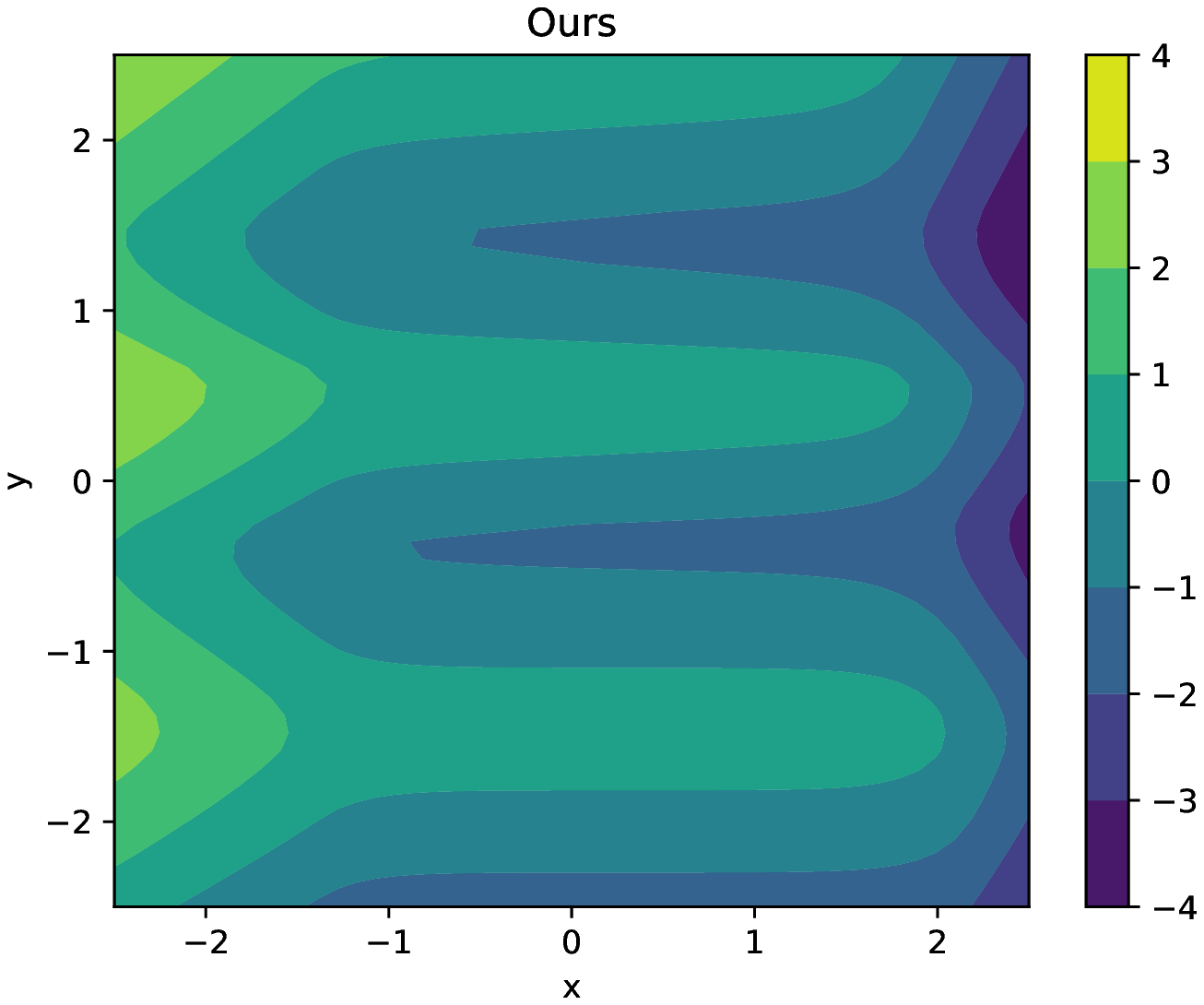}
        \caption{}
        \label{subfig:results_ours}
    \end{subfigure}
    \hfill
    
    \caption{Results for function fitting experiments where (a)  is the ground truth, (b) is the result obtained from a simple unconstrained neural network (with ELU activation) and (c) is the result obtained from a constrained neural network (weights to be positive and ELU activation) and (d) shows our results with constrained weights and bipolar ELU activation}
    \label{fig:appendix_results}
\end{figure*}

Apart from the experimental results in the main part of the paper, here we highlight the usefulness of the our monotonic dense unit. In a scenario where the dataset is noisy and very small and also contains monotonic features, our neural network constructed using our proposed monotonic dense unit tends to perform better because of the inductive bias. To illustrate this we consider a synthetic dataset generated by function $f(x,y) =  sgn(ax)~x^3 + b~\sin(cy), a,b,c\in\{0.5, 0.35, 3.3\}$. We sample points uniformly in the range $(-2.5, 2.5)$ for both $x$ and $y$, and add zero-centered Gaussian noise. We then sample only $100$ points for training and train three neural networks having the same kind of architecture but having difference in constraints on weights as mentioned in section 3, but activation function used is ELU instead of ReLU. We test the three networks - Unconstrained, Constrained ELU and Constrained bipolar ELU, to on the task of fitting the aforementioned function. The results be seen in the Figure. \ref{fig:appendix_results}. It is evident that the unconstrained neural network does not preserve monotonicity and although constrained neural network does preserve monotonicity, it cannot faithfully reproduce the concave part of the function, whereas ours (constrained bipolar ELU) can fit both concave and convex part of the function.  

\bibliography{references}